\documentclass{isrpaper}

\ifcsname XeTeXrevision\endcsname\usepackage{fontspec}\else\usepackage{lmodern}\fi

\usepackage{orcidlink}
\extrafloats{100}
\usepackage{array}
\definecolor{RoyalBlue}{HTML}{1D5FA7}
\definecolor{Teal}{HTML}{0F766E}
\definecolor{Crimson}{HTML}{B83227}

\AtBeginDocument{\setlength{\abovedisplayskip}{5pt plus 1pt minus 1pt}\setlength{\belowdisplayskip}{5pt plus 1pt minus 1pt}}

\usepackage[
    backend=biber,
    style=authoryear,
    natbib=true,
    doi=false,
    url=false,
    isbn=false,
    maxbibnames=3,
    minbibnames=1,
    maxcitenames=2,
    uniquelist=false,
]{biblatex}

\AtEveryBibitem{%
   \clearfield{Title}%
   \clearfield{howpublished}%
   \clearfield{eprint}%
   \clearfield{note}%
}

\renewcommand{\citet}[1]{\textcite{#1}}
\renewcommand{\citep}[1]{\parencite{#1}}

\title{FASCIA: Seeing Beneath the Surface with Queryable, Personalized Anatomy in Any Pose}

\author{
R. James Cotton\,\orcidlink{0000-0001-5714-1400}\,$^{1,2}$, Tim Unger$^{3,4}$, Lucinda Williamson$^{1}$}

\affiliations{
$^{1}$Shirley Ryan AbilityLab \\ $^{2}$Department of Physical Medicine and Rehabilitation, Northwestern University \\ $^{3}$Rehabilitation Engineering Laboratory, Department of Health Sciences and Technology, ETH Zurich, Zurich, Switzerland \\ $^{4}$Therapy Science Lab, Lake Lucerne Institute (LLUI), Vitznau, Switzerland}

\paperdate{November 2026}

\correspondence{rcotton@sralab.org}

\paperabstract{We present FASCIA (\textit{Functional Anatomical Structures and Composition via Implicit Atlas}), a Momentum Human Rig (MHR)-anchored framework that predicts and poses a person's internal anatomy from their external body shape and skeletal proportions. FASCIA maps a single 290-structure population template onto each individual through a smooth, invertible warp. As people with the same shape can differ internally, a body-dependent Gaussian prior predicts both average warp coordinates and their inter-subject variability. The warp deforms a learned base template into bones, muscles, organs, and fat depots. Extending MHR skinning into the interior enables direct supervision of this model across four automatically labeled CT cohorts and one whole-body MRI cohort. On 82 held-out CT scans, FASCIA reaches a mean volumetric Dice (volume agreement) of 0.48 on named bones and 0.63 on muscles, keeps the majority of bones and muscles in one piece, and yields calibrated voxel probabilities. On 33 held-out whole-body MRI subjects, FASCIA's pointwise field improves tissue overlap over HIT across all compartments, predicts between-person differences in whole-body adipose percentage ($R^2 = 0.88$ vs. 0.38 in women and 0.54 vs. -0.01 in men), and recovers segment centers of mass within 1.1 mm. Because it needs only an MHR fit, FASCIA can be driven from images, video, or any system that outputs MHR. This opens the door to fitting individualized calibrated musculoskeletal digital twins to video motion data.

}

\begin{document}

\maketitle

\section{Introduction}

Markerless motion capture measures clinical kinematics from multiview and monocular video \parencite{kanko_concurrent_2021, uhlrich2023opencap, cotton_optimizing_2023, cimorelli_validation_2024, peiffer_portable_2026, cotton_biomechanical_2026}, yet computing accurate joint moments, muscle forces, and tissue-level biomarkers requires knowing how mass and tissue are distributed beneath the skin. Standard musculoskeletal pipelines \parencite{delp_opensim_2007, werling2023addbiomechanics} scale generic template models by segment length and initialize segment masses and inertias from template defaults---often derived from male cohorts---or regression tables built on small historical samples \parencite{deleva1996adjustments, dumas2007adjustments}, whereas subject-specific inertias capture true body-composition variability and alter inverse-dynamics joint moments \parencite{akhundov2022subject, gambietz2026from}. With rapid advances in mesh-based human recovery from images and video, learning from medical imaging where internal tissues lie relative to a body mesh opens a direct path to more accurate biomechanical analysis grounded in modern computer vision.

A growing body of work predicts internal anatomy from the outer surface. Parametric hulls such as SMPL \parencite{loper_smpl_2015} enclose total volume, and OSSO \parencite{keller2022osso}, SKEL \parencite{keller2023skel}, and BOSS \parencite{shetty2023boss} place discrete bones and organs inside the skin. HIT \parencite{keller2024hit}, the closest prior work, classifies points inside an SMPL mesh from whole-body magnetic resonance imaging (MRI) into bone, subcutaneous fat, lean tissue merged with visceral fat, or empty space, but compresses predicted tissue volumes toward the cohort mean rather than capturing individual variability. In addition, the HIT MRI dataset provides only five coarse tissue masks without raw scans for relabeling. Alternatively, public computed tomography (CT) datasets \parencite{jimenezdeltoro2016visceral, gutschmayer2025quadrahc, fischer2023vsd, hinge2026copenhagen} processed with automated segmentation methods \parencite{wasserthal2023totalsegmentator, sundar2022moose, mckay2024musclemap, xu2025cads} can resolve labels for individual bones, muscles, and organs. Leveraging these datasets, however, introduces a key supervision challenge because individual scans frequently provide only partial coverage of the body and each automated segmenter labels only its own subset of structures.

\begin{figure}[!htb]
\centering
\includegraphics[width=0.99\linewidth,height=0.38\textheight,keepaspectratio]{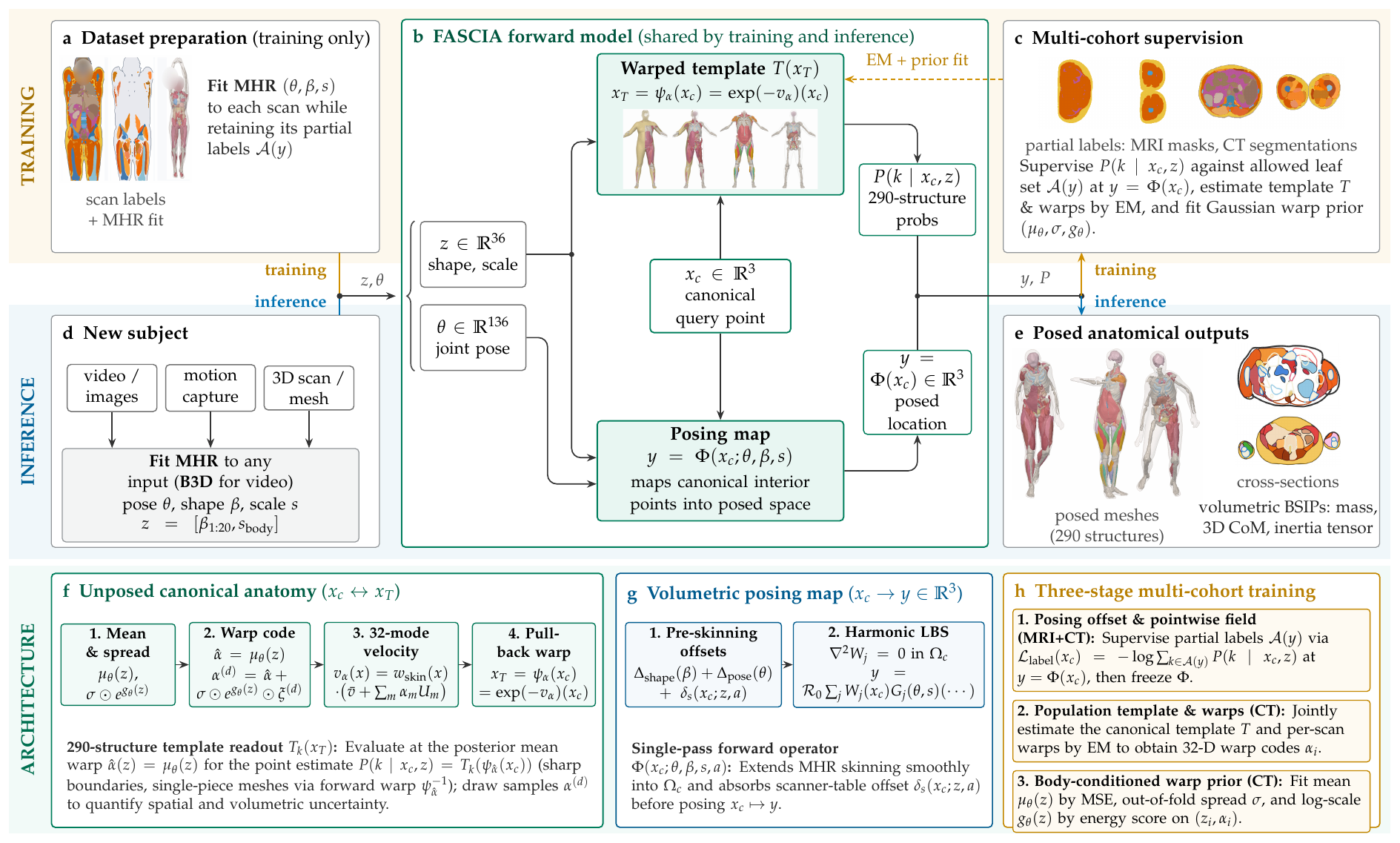}
\caption[]{\textbf{FASCIA overview and architecture.} \textbf{Top (a--e):} \textbf{(a)} Fit MHR $(\theta, \beta, s)$ to each training scan while retaining its partial labels $\mathcal{A}(y)$. \textbf{(b)} Canonical $x_c$, body code $z$, and pose $\theta$ yield 290-structure probabilities $P(k \mid x_c, z)$ and posed location $y = \Phi(x_c)$. \textbf{(c)} Partial-label supervision estimates template $T$, per-scan warps, and warp prior $(\mu_\theta, \sigma, g_\theta)$. \textbf{(d, e)} Fitting MHR to any input (video, motion capture, or 3D scan) drives posed anatomical meshes, cross-sections, and BSIPs. \textbf{Bottom (f--h):} \textbf{(f)} Gaussian warp prior maps $z$ to mean $\hat{\alpha}$ and draws $\alpha^{(d)}$ driving velocity $v_\alpha$ and pullback $x_T = \psi_\alpha(x_c)$ into $T_k(x_T)$. \textbf{(g)} Forward posing map $y = \Phi(x_c)$ combines MHR interior skinning and blendshapes with scanner-table offset $\delta_s$. \textbf{(h)} Three-stage multi-cohort training.}
\label{fig-overview}
\end{figure}

We introduce FASCIA (\textit{Functional Anatomical Structures and Composition via Implicit Atlas}), anchored to the Momentum Human Rig (MHR) \parencite{ferguson_mhr_2025} (Figure~\ref{fig-overview}), which supplies joint angles $\theta$, body shape $\beta$, and bone scales $s$, from which we form a compact shape-and-scale code $z$. In canonical rest space, FASCIA deforms a 290-structure population template onto each person through a smooth, invertible 3D warp $\psi_\alpha$. This warp morphs bone and soft-tissue geometry while keeping thin curved bones such as ribs, scapulae, and clavicles in one piece where a pointwise neural field fragments them. Because people with the same exterior still differ internally, a body-dependent Gaussian prior predicts both the mean warp and variability around it for spatial and volumetric uncertainty. Extending MHR's skinning into the 3D interior forms a forward posing map $\Phi$ that pushes canonical points into posed CT and MRI scans to supervise training against the extracted tissue labels.

Our contributions are:

\begin{enumerate}
\item A deformable anatomical template and body-dependent Gaussian warp prior conditioned on MHR shape and scale that predict individualized, posed anatomy with calibrated voxel probabilities and sample-based spatial and volumetric uncertainties.
\item A forward volumetric posing map and partial-label objective trained jointly across five heterogeneously labeled CT and MRI datasets without unposing scans, paired with closed-form mass calibration for personalized body-segment inertial parameters (BSIPs).
\item Empirical validation showing that FASCIA captures individual differences in MRI body composition where HIT compresses predictions toward the cohort mean, while an identically conditioned pointwise CT ablation demonstrates that template warping keeps thin curved bones whole and improves segmentation accuracy.
\item Driving personalized internal anatomy directly from images or monocular video through any system that outputs MHR, with fast GPU inference.
\end{enumerate}

\section{Related Work}

Computational anatomy models population structure via an unbiased template deformed onto each individual by a diffeomorphic warp \parencite{joshi2004unbiased, ashburner2007dartel}, with networks predicting conditional deformations in one pass \parencite{dalca2019conditional}; FASCIA embeds this template-and-warp formulation inside an articulated body rig. Outer-surface rigs such as SMPL \parencite{loper_smpl_2015}, Anny \parencite{bregier2025anny}, and MHR \parencite{ferguson_mhr_2025} parameterize external shape and linear blend skinning. MHR uses a 127-joint skeleton and explicit bone scales, and B3D \parencite{cotton_biomechanical_2026} jointly recovers the MHR skin surface, skeletal kinematics, and bone scales from images and video.

Internal-anatomy regressors typically predict discrete shells: OSSO \parencite{keller2022osso} learns from dual-energy X-ray absorptiometry (DXA) scans to predict a skeleton from STAR shape, then optimizes its pose inside the body under anatomical constraints; SKEL \parencite{keller2023skel} rigs biomechanical joints; BOSS \parencite{shetty2023boss} adds major organs; and SOMA \parencite{alvarado2026soma} drives person-specific muscle meshes from pose. However, discrete shells leave surrounding soft tissue unpartitioned, shape regressors \parencite{qiao2024bodycomp} predict only scalar total and regional summaries, and hull-based inertia estimators \parencite{gambietz2026from} rely on HIT's coarse compartments even though personalized tissue distribution alters inverse-dynamics moments \parencite{akhundov2022subject}. HIT \parencite{keller2024hit} classifies interior SMPL points on whole-body MRI into three coarse tissues; FASCIA extends volumetric modeling to individual bones, muscles, organs, and fat depots via diffeomorphic template warping and forward volumetric skinning.

\section{Methods}

FASCIA predicts 3D internal anatomy across $K$ anatomical classes from MHR \parencite{ferguson_mhr_2025} shape $\beta$, bone scales $s$, and pose $\theta$ (Figure~\ref{fig-overview}). It factorizes internal 3D anatomy into two forward maps (Figure~\ref{fig-overview}b, f, g): for each canonical point $x_c \in \Omega_c \subset \mathbb{R}^3$, a smooth invertible warp $x_T = \psi_\alpha(x_c)$ with subject weights $\alpha$ deforms a shared template $T(x_T)$ conditioned on a shape-and-scale code $z$. A forward posing map $y = \Phi(x_c) \in \mathbb{R}^3$ then pushes canonical points into posed scans across five heterogeneously labeled CT and MRI cohorts (Figure~\ref{fig-overview}a, c, h) and poses volumes, meshes, and inertias at inference (Figure~\ref{fig-overview}d, e).

\subsection{Datasets, Anatomical Hierarchy, and Registration}

We train on 584 scans (252 MRI, 332 CT) and evaluate on 115 subject-disjoint test scans (33 MRI, 82 CT; Table~\ref{tab-datasets}). On the HIT whole-body MRI benchmark \parencite{keller2024hit}, we use the five released voxelwise tissue labels directly, whereas the four CT cohorts are segmented with four automated tools covering complementary structures: TotalSegmentator \parencite{wasserthal2023totalsegmentator}, MuscleMap \parencite{mckay2024musclemap}, MOOSE \parencite{sundar2022moose}, and CADS \parencite{xu2025cads}. Manual VSD lower-limb bone labels were reserved for held-out evaluation. Labels map into a unified hierarchy of $K = 290$ anatomical structures (79 scored named bones, with each foot's calcaneus, talus, and mid-foot tarsals as separate classes; 108 named trunk, pelvic, shoulder, neck, and lower-limb muscles; and organs, vessels, fat depots, and regional soft tissues; arm muscles remain unparcellated because MuscleMap does not segment upper-limb muscles), organized into coarse tissue groups matching HIT's five MRI classes.

\begin{table}[!htb]
\centering\scriptsize\setlength{\tabcolsep}{2.8pt}\renewcommand{\arraystretch}{1.02}
\centering
\caption[]{Subject-disjoint training and test cohorts. \textsuperscript{a}Official 317/41/40 splits after registration quality control.}
\label{tab-datasets}
\begin{tabular}{>{\raggedright\arraybackslash}p{\dimexpr 0.23\linewidth-2\tabcolsep}>{\centering\arraybackslash}p{\dimexpr 0.19\linewidth-2\tabcolsep}>{\raggedright\arraybackslash}p{\dimexpr 0.19\linewidth-2\tabcolsep}>{\raggedright\arraybackslash}p{\dimexpr 0.39\linewidth-2\tabcolsep}}
\toprule
Cohort & Scans (train / test) & Coverage, posture & Labels \\
\hline
HIT whole-body MRI \parencite{keller2024hit} & 398 scans (252 / 33 subj.)\textsuperscript{a} & Neck to ankle, prone & HIT's 5 tissue classes \\
VISCERAL CT \parencite{jimenezdeltoro2016visceral} & 154 (123 / 31) & Whole body or trunk, supine & CADS, TotalSegmentator, MuscleMap \\
QUADRA PET/CT \parencite{gutschmayer2025quadrahc} & 96 (78 / 18) & Total body, supine, repeat scans & MOOSE, TotalSegmentator, MuscleMap \\
VSD CT \parencite{fischer2023vsd} & 84 (67 / 17) & Whole body or lower limb, postmortem & TotalSegmentator, MuscleMap; manual lower-limb bones \\
Copenhagen PET/CT \parencite{hinge2026copenhagen} & 80 (64 / 16) & Total body, supine & TotalSegmentator, MuscleMap \\
\bottomrule
\end{tabular}
\end{table}

We register MHR \parencite{ferguson_mhr_2025}, which is parameterized by joint angles $\theta$, shape $\beta$, bone scales $s$, and global placement $\mathcal{R}_0 \in \mathrm{SE}(3)$ to each scan. On CT, we estimate these parameters by nonlinear least squares fitted strictly to the outer skin contour and segmented bones without reading soft-tissue labels, masking misregistered limbs and imputing out-of-field bone scales from a pretrained body-shape prior; on HIT MRI, we fit MHR directly to the released SMPL \parencite{loper_smpl_2015} meshes. Each subject's internal anatomy is conditioned on the body code $z = [\beta_{1:20}, s_{\text{body}}] \in \mathbb{R}^{36}$, concatenating the leading 20 shape coefficients and 16 non-hand bone scales.

\subsection{The FASCIA Forward Model}

Standard MHR \parencite{ferguson_mhr_2025} deforms only its outer skin mesh via surface blendshapes and rigid joint frames $G_j(\theta, s) \in \mathrm{SE}(3)$. FASCIA pairs canonical personalization---warping $x_c \in \Omega_c$ into template space $\Omega_T \subset \mathbb{R}^3$ via $x_T = \psi_\alpha(x_c)$ to read class-$k$ probabilities $P(k \mid x_c, z) = T_k(x_T)$---with forward volumetric posing $y = \Phi(x_c; \theta, \beta, s) \in \mathbb{R}^3$. Across both stages (Figure~\ref{fig-overview}f, g), MHR supplies only the outer skin boundary, joint tree, scales $s$, and frames $G_j(\theta, s)$. All internal bone and soft-tissue geometries originate in the canonical template $T$ and morph non-rigidly in rest space via $\psi_\alpha$ while preserving connectivity; under posture $\theta$, $\Phi$ blends joint transforms smoothly across interior voxels so soft tissues move coherently with the skeleton, whereas exported cortical bone meshes attach rigidly to $G_j(\theta, s)$.

\subsubsection{Canonical Template and Diffeomorphic Warp Prior}

Our ablation shows that pointwise voxel classification from the body code blurs thin structures across subjects into fragments. Instead, FASCIA learns a shared template $T(x_T) \in [0, 1]^{K}$ on a $5\text{ mm}$ canonical grid ($\Omega_T \subset \mathbb{R}^3$). This is deformed with a stationary velocity field $v_\alpha(x)$ parameterized on a $15\text{ mm}$ lattice by weights $\alpha \in \mathbb{R}^{32}$ over orthonormal modes $U_{1:32}(x)$ around mean velocity $\bar{v}(x)$ \parencite{ashburner2007dartel}:

\begin{equation}
\label{eq-flow}
\begin{aligned}
v_\alpha(x) &= w_{\text{skin}}(x)\!\left(\bar{v}(x) + \sum_{m=1}^{32} \alpha_m\, U_m(x)\right), \\
\psi_\alpha &= \exp(-v_\alpha) : \Omega_c \to \Omega_T, \\
\psi_\alpha^{-1} &= \exp(+v_\alpha) : \Omega_T \to \Omega_c
\end{aligned}
\end{equation}

Here $w_{\text{skin}}(x) \in [0, 1]$ constrains the velocity field near the canonical skin, slowing it smoothly to zero so $\psi_\alpha$ leaves the canonical outer skin boundary fixed. Integrating $\pm v_\alpha$ via six scaling-and-squaring steps \parencite{ashburner2007dartel} gives the pullback diffeomorphism $\psi_\alpha$ mapping $x_c$ to $x_T$ for probability lookup and its exact inverse $\psi_\alpha^{ -1}$ pushing template meshes into $\Omega_c$.

Because external shape $z$ does not uniquely fix internal boundaries, FASCIA predicts and samples $\alpha$ under a diagonal Gaussian prior (Figure~\ref{fig-overview}f):

\begin{equation}
\label{eq-flow-draw}
\alpha^{(d)} = \mu_\theta(z) + \sigma \odot \exp\!\big(g_\theta(z)\big) \odot \xi^{(d)}, \qquad \xi^{(d)} \sim \mathcal{N}(0, I_{32})
\end{equation}

Here the neural network $\mu_\theta(z)$ (with learned parameters distinct from pose $\theta$) predicts the expected warp $\hat{\alpha}(z) = \mu_\theta(z)$, while baseline mode scales $\sigma \in \mathbb{R}_+^{32}$ and log-spread network $g_\theta(z)$ form a noise model ($\odot$: elementwise product) driving reparameterized draws \parencite{kingma2014autoencoding}. Evaluating $T$ through a single warp $\psi_\alpha$ keeps structures whole:

\begin{equation}
\label{eq-warp-modes}
P(k \mid x_c, \alpha) = T_k\!\big(\psi_{\alpha}(x_c)\big), \qquad \alpha = \hat{\alpha}(z) \text{ or } \alpha^{(d)}
\end{equation}

Point predictions and meshes evaluate $\hat{\alpha}(z)$, while evaluating across posterior draws $\alpha^{(d)}$ yields empirical confidence intervals on centroids and volumes without blurring.

\subsubsection{Volumetric Posing Map}

To pose $x_c \in \Omega_c$ to $y \in \mathbb{R}^3$ (Figure~\ref{fig-overview}g), FASCIA extends MHR skinning and blendshapes into the interior via Laplace's equation ($\nabla^2 u = 0$) with MHR's skin values as boundary conditions, plus a learned pre-skinning displacement $\delta_s(x_c; z, a)$ with prone-MRI indicator $a \in \{0, 1\}$ absorbing habitus tissue shifts and scanner-table compression before joint rotations:

\begin{equation}
\label{eq-phi}
\begin{aligned}
&y = \Phi(x_c; \theta, \beta, s, a) = \mathcal{R}_0\Bigg( \sum_{j=1}^{J} \textcolor{Teal}{\underbrace{W_j(x_c)}_{\substack{\text{FASCIA fixed} \\ \text{harmonic weights}}}}\, \textcolor{RoyalBlue}{\underbrace{G_j(\theta, s)}_{\substack{\text{from MHR:} \\ \text{joint frames}}}} \\
&\;\cdot \Big( x_c + \textcolor{Teal}{\underbrace{\Delta_{\text{shape}}(x_c; \beta) + \Delta_{\text{pose}}(x_c; \theta)}_{\substack{\text{FASCIA fixed} \\ \text{harmonic blendshapes}}}} + \textcolor{Crimson}{\underbrace{\delta_s(x_c; z, a)}_{\substack{\text{FASCIA learned} \\ \text{tissue/table shift}}}} \Big) \Bigg)
\end{aligned}
\end{equation}

The MHR model provides the rigid joint frames $G_j(\theta, s)$, and outer-skin boundary values, while FASCIA extends those surface weights and blendshapes into the 3D interior as parameter-free harmonic fields: $W_j$ and $\Delta_{\text{shape}}$ are solved once and stored on an $8\text{ mm}$ lattice, and MHR's pose correctives are extended into $\Delta_{\text{pose}}$ by the same harmonic solve at each pose. For the shape blendshapes $\Delta_{\text{shape}}$, canonical bone nodes act as depth-damped interior Dirichlet boundaries ($\exp( -(d_{\text{skin}}/20\text{ mm})^2)$, $\sigma = 20\text{ mm}$) so deep bones remain anchored to the skeletal scale $s$ rather than drifting with adiposity---in contrast to HIT's unanchored volumetric blendshapes---while superficial bones follow the overlying skin so thin subcutaneous layers never invert. Like HIT, FASCIA accounts for tissue deformations from laying flat on a scanner. Specifically, we learn $\delta_s(x_c; z, a) = \sum_n N_n(x_c)(D_n z + a\, e_n)$ from scans on a $24\text{ mm}$ cubic B-spline lattice $N_n(x_c)$, where $D_n \in \mathbb{R}^{3 \times 36}$ models body-habitus shifts and $e_n \in \mathbb{R}^3$ absorbs prone MRI table compression ($a = 1$ prone MRI; $a = 0$ supine CT and inference). To monitor model stability, $\det\nabla_{x_c}\Phi$ measures local volume change and $\det\nabla_{x_c}\Phi \le 0$ flags implausible tissue folds, which are penalized in training and monitored under reposing in Results. As a smooth forward map, $\Phi$ pushes canonical samples $x_c$ directly into posed scans to read labels at $y = \Phi(x_c)$ for training (Figure~\ref{fig-overview}c).

\subsection{Training Objective and Optimization}

\textbf{Partial-label supervision.} Most training scans include only a partial body, so for scan $i$ we sample canonical points $x_c \sim q(x_c)$ from an interior proposal density $q$ that defines the canonical interior region of the body to be supervised, and push them forward to $y_i = \Phi_i(x_c) \equiv \Phi(x_c; \theta_i, \beta_i, s_i, a_i)$ (Figure~\ref{fig-overview}c). Each segmentation source specifies an allowed class set $\mathcal{A}_i(y_i) \subseteq \{1, \dots, K\}$: one leaf for a fine CT label such as the femur, or all leaves inside a coarse label such as HIT's broad muscle category. Weighting by posed volume $w_i(x_c) = |\det\nabla_{x_c}\Phi_i(x_c)| / q(x_c)$ integrates the negative log-likelihood of scan $i$'s predicted class distribution $P_i(k \mid x_c)$ over scan volume:

\begin{equation}
\label{eq-groupsum}
\begin{aligned}
\mathcal{L}_{\text{label}}(i) &= \frac{1}{\sum_{x_c} w_i(x_c)} \sum_{x_c \sim q} \underbrace{w_i(x_c)}_{\substack{\text{posed vol.} \\ |\det\nabla\Phi_i|/q}} \\
&\quad \cdot \Bigg(\! -\log \!\!\underbrace{\sum_{k \in \mathcal{A}_i(y_i)} P_i(k \mid x_c)}_{\substack{\text{allowed-leaf mass} \\ \text{at } y_i = \Phi(x_c;\,\theta_i,\beta_i,s_i,a_i)}} \!\!\Bigg)
\end{aligned}
\end{equation}

For one leaf, \textbf{Eq.~\ref{eq-groupsum}} is cross-entropy; for a coarse class, it supervises compartment mass while CT resolves fine leaves.

\textbf{Learning the posing offset and canonical template.} Training proceeds in three sequential steps (Figure~\ref{fig-overview}h) because prone table compression $e_n$ is observed only on coarse-labeled MRI, template estimation requires a fixed canonical-to-scan map $\Phi$ on finely labeled CT, and test-time inference observes only $z$. First, across the full MRI and CT training set, we jointly fit $\delta_s$ and a continuous pointwise neural field $f(x_c, z)$ predicting class probabilities directly from $(x_c, z)$ under \textbf{Eq.~\ref{eq-groupsum}} and then freeze $\Phi$. Second, with $\Phi$ fixed, we estimate the canonical template $T$ and velocity basis $U_{1:32}$ across the CT training cohort by coarse-to-fine expectation--maximization: each E-step fits a Gaussian-regularized per-scan velocity field to that scan's CT labels under the unweighted allowed-leaf log-likelihood of \textbf{Eq.~\ref{eq-groupsum}} on the canonical lattice (re-centering $\bar{v} = \mathbf{0}$), and each M-step updates $T$ from the warped label responsibilities. Connectivity cleanup and singular value decomposition of the fitted velocity fields then yield the orthonormal modes $U_{1:32}$ and each CT training scan's label-fitted warp code $\alpha_i \in \mathbb{R}^{32}$.

\textbf{Learning the body-conditioned warp prior.} Because $\alpha_i$ in Step 2 is fitted to CT labels that are unavailable for unseen subjects, Step 3 trains the prior networks $\mu_\theta$ and $g_\theta$ in \textbf{Eq.~\ref{eq-flow-draw}} on the paired CT training codes $(z_i, \alpha_i)$ to predict warps from outer body shape and scale alone. We fit the mean network $\mu_\theta(z)$ by regularized mean-squared error against $\alpha_i$. Using five-fold cross-validation across the training scans, we set $\sigma$ to the per-mode root-mean-square error of out-of-fold residuals $r_i = \alpha_i - \mu_\theta^{\text{oof}}(z_i)$---where $\mu_\theta^{\text{oof}}(z_i)$ excludes scan $i$'s fold---and train the spread network $g_\theta(z)$ by minimizing the multivariate energy score \parencite{gneiting2007strictly} on $r_i$.

\subsection{Posed Outputs, Inertial Parameters, and Ablations}

\textbf{Posed 3D meshes and BSIPs.} Marching-cubes meshes from $T$ map into rest space by $\psi_{\hat{\alpha}(z)}^{ -1}$; under $\theta$, soft tissues deform via $\Phi$ while cortical bone meshes attach rigidly to $G_j(\theta, s)$. Integrating effective density $\rho(y) = \sum_k \rho_k P(k \mid x_c, z)$ over per-class literature densities $\rho_k$ \parencite{gambietz2026from} yields personalized segment masses, 3D centers of mass, and inertia tensors \parencite{dumas2018estimation}. Given a measured body mass $m_{\text{body}}$, two-compartment mass calibration (Archimedes balance) reconciles the integrated mass $\hat{m}_{\text{body}}$ at fixed hull volume by trading subcutaneous adipose ($\text{AT}$) against lean tissue ($\text{LT}$) volume, $\Delta V_{\text{AT}} = -\Delta V_{\text{LT}} = (m_{\text{body}} - \hat{m}_{\text{body}})/(\rho_{\text{AT}} - \rho_{\text{LT}})$.

\textbf{Ablations.} To isolate diffeomorphic template warping, we compare against three baselines sharing the same forward posing map $\Phi$: (1) \textbf{Pointwise neural field $f(x_c, z)$}, which omits the template and warp to classify canonical coordinates directly from $z$ and, because Step 1 trains $f$ jointly on MRI and CT whereas Step 2 estimates $T$ on CT alone, serves as our MRI coarse-tissue predictor; (2) \textbf{Zero warp}, which evaluates the unwarped template $T(x_c)$ at $\alpha = \mathbf{0}$; and (3) \textbf{Oracle warp}, an upper bound that fits $\alpha$ directly to held-out scan labels with $T$ fixed.

\textbf{Evaluation metrics.} To quantify spatial accuracy and topological plausibility across structures spanning three orders of magnitude in volume, we report two complementary metrics across held-out cohorts: \textbf{volumetric Dice}, $\text{Dice}(A, B) = 2|A \cap B| / (|A| + |B|) \in [0, 1]$, which normalizes overlap by structure size across small bones, thin muscles, and large visceral organs, and the \textbf{single-piece rate} ($1 - \text{fragmentation}$), the fraction of predicted structure instances forming a single connected 3D component.

\section{Results}

\subsection{Qualitative Results}

\begin{figure}[!htb]
\centering
\includegraphics[width=0.99\linewidth,height=0.40\textheight,keepaspectratio]{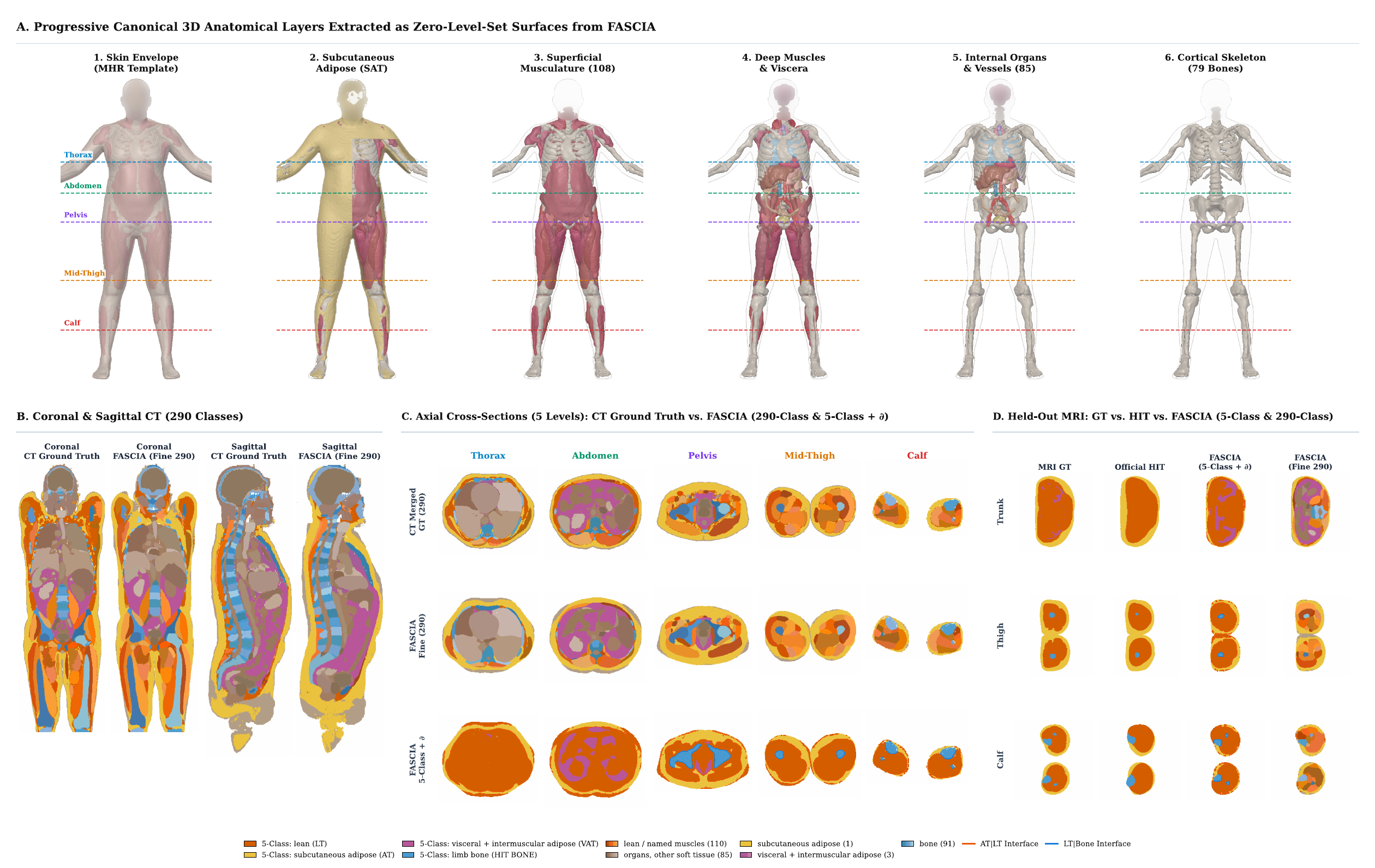}
\caption[]{\textbf{Predicted anatomy on held-out subjects.} \textbf{(A)} Six nested canonical layers carried by the FASCIA mean warp. \textbf{(B)} Coronal and sagittal CT reference versus FASCIA cross-sections. \textbf{(C)} Axial CT sections across fine anatomical structures and coarse tissue groups with level-set interfaces ($\partial$). \textbf{(D)} Held-out MRI sections comparing ground truth, HIT \parencite{keller2024hit}, and FASCIA (coarse tissues and full anatomy).}
\label{fig-layers}
\end{figure}

We evaluate FASCIA across the held-out MRI and CT test cohorts by fitting only the outer MHR surface without access to test-time tissue labels. Conditioned on $z$ alone, FASCIA reconstructs nested 3D layers from skin through subcutaneous fat, skeletal muscles, visceral organs, and the cortical skeleton (Figure~\ref{fig-layers}A), aligns with coronal, sagittal, and axial CT cross-sections across both fine structures and coarse tissues (Figure~\ref{fig-layers}B,C), and captures individual adipose and lean distributions on MRI (Figure~\ref{fig-layers}D).

\subsection{Whole-Body MRI Tissue Composition and Segment Inertias}

\begin{figure}[!htb]
\centering
\includegraphics[width=0.99\linewidth,height=0.37\textheight,keepaspectratio]{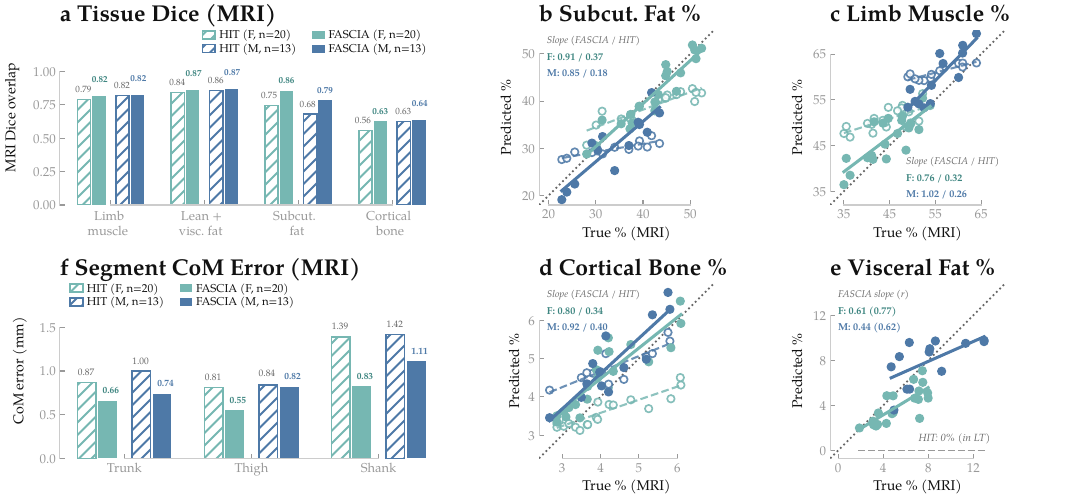}
\caption[]{\textbf{Whole-body MRI tissue composition and segment inertial parameters.} \textbf{(a)} Held-out MRI Dice ($n=33$) comparing FASCIA and HIT \parencite{keller2024hit} by sex on appendicular \texttt{Limb muscle}, \texttt{Lean + visc. fat}, subcutaneous fat, and cortical bone. \textbf{(b--e)} Predicted versus true volume percentages for subcutaneous fat \textbf{(b)}, limb muscle \textbf{(c)}, cortical bone \textbf{(d)}, and visceral fat \textbf{(e)} (HIT: $0\%$, merged in \texttt{LT}); open circles and dashed lines denote HIT, filled circles and solid lines denote FASCIA, colors match \textbf{(a, f)} (teal: female; blue: male), dotted lines mark $1{:}1$ identity, and insets report per-sex slopes. \textbf{(f)} Segment center-of-mass 3D location error (mm) on trunk, thigh, and shank comparing FASCIA and HIT by sex.}
\label{fig-e1-bsips}
\end{figure}

We first benchmark FASCIA against the official pretrained HIT \parencite{keller2024hit} on the held-out whole-body MRI cohort (Figure~\ref{fig-e1-bsips}, Table~\ref{tab-e1}). Although the HIT dataset releases separate visceral fat (\texttt{VAT}) and lean tissue (\texttt{LT}) masks, the trained HIT network merges \texttt{VAT} into \texttt{LT} (\texttt{LT+VAT})---making \texttt{LT} within the visceral-free arms and legs a direct measure of appendicular limb muscle, alongside whole-body \texttt{LT+VAT}, subcutaneous fat, and cortical bone; FASCIA improves volumetric Dice on every compartment (outperforming HIT on subcutaneous fat for every held-out subject) while resolving visceral fat separately (Figure~\ref{fig-e1-bsips}a,e, Table~\ref{tab-e1}).

\begin{table}[!htb]
\centering\scriptsize\setlength{\tabcolsep}{2.8pt}\renewcommand{\arraystretch}{1.02}
\centering
\caption[]{Held-out whole-body MRI benchmark ($n=33$: 20 female, 13 male). Voxel Dice is reported for lean tissue plus visceral fat (LT+VAT), subcutaneous adipose tissue (AT), and cortical bone (HIT's three non-background classes; FASCIA also resolves visceral adipose tissue separately). On MRI, FASCIA is read through its pointwise field $f(x_c, z)$ (three seeds, averaged per subject). Adipose \% $R^2$ evaluates whole-body adipose fraction; SD ratio is the predicted over the true cohort standard deviation (1.00 means the predicted inter-subject spread matches the ground-truth spread). CoM and MOI give the range of segment center-of-mass error (mm) and principal moment-of-inertia error (\%) across trunk, thigh, and shank in both sexes. Mass calibration rescales regional volumes to the reference hull and applies the Archimedes mass-balance update.}
\label{tab-e1}
\begin{tabular}{>{\raggedright\arraybackslash}p{\dimexpr 0.23\linewidth-2\tabcolsep}>{\centering\arraybackslash}p{\dimexpr 0.14\linewidth-2\tabcolsep}>{\centering\arraybackslash}p{\dimexpr 0.14\linewidth-2\tabcolsep}>{\centering\arraybackslash}p{\dimexpr 0.14\linewidth-2\tabcolsep}>{\centering\arraybackslash}p{\dimexpr 0.12\linewidth-2\tabcolsep}>{\centering\arraybackslash}p{\dimexpr 0.11\linewidth-2\tabcolsep}>{\centering\arraybackslash}p{\dimexpr 0.12\linewidth-2\tabcolsep}}
\toprule
Method & LT+VAT / AT / Bone Dice (F) & LT+VAT / AT / Bone Dice (M) & Adipose \% $R^2$ (F / M) & SD ratio (F / M) & CoM (mm) & MOI (\%) \\
\hline
Anthropometric tables \parencite{dumas2007adjustments} & --- & --- & --- & --- & 12.5--45.6 & 13.5--38.5 \\
Uniform-density hull & --- & --- & --- & --- & 0.73--1.60 & 1.60--4.70 \\
Stature + mass regression & --- & --- & 0.86 / 0.75 & 0.81 / 0.76 & --- & --- \\
HIT \parencite{keller2024hit} & 0.844 / 0.746 / 0.557 & 0.859 / 0.683 / 0.627 & 0.38 / -0.01 & 0.42 / 0.20 & 0.81--1.42 & 0.77--3.12 \\
FASCIA pointwise field $f(x_c, z)$ & \textbf{0.865} / \textbf{0.857} / \textbf{0.629} & \textbf{0.873} / \textbf{0.787} / \textbf{0.640} & 0.88 / 0.54 & 0.96 / 0.96 & 0.55--1.11 & 0.49--2.23 \\
FASCIA pointwise field + mass calibration & \textbf{0.865} / \textbf{0.857} / \textbf{0.629} & \textbf{0.873} / \textbf{0.787} / \textbf{0.640} & 0.89 / 0.87 & 1.06 / 0.90 & 0.55--1.11 & 0.34--1.62 \\
\bottomrule
\end{tabular}
\end{table}

Beyond mean voxel overlap, precise biomechanical modeling requires capturing inter-subject anatomical differences rather than only cohort-mean overlap (Figure~\ref{fig-e1-bsips}b--e, Table~\ref{tab-e1}). Across subcutaneous fat, limb muscle, and cortical bone, HIT compresses predictions toward the cohort mean, whereas FASCIA shows slopes closer to identity (Figure~\ref{fig-e1-bsips}b--e). From outer surface shape alone, FASCIA outperforms HIT, constant-density hulls, and anthropometric regression on both whole-body fat-percentage variance ($R^2 = 0.88$ in women and 0.54 in men) and 3D segment inertial parameters, and adding body-mass calibration removes a small male fat-versus-lean offset ($R^2 = 0.87$; Table~\ref{tab-e1}). Integrating literature tissue densities over these personalized compartments captures inter-subject variation in segment density where uniform hulls cannot and lowers segment center-of-mass (Figure~\ref{fig-e1-bsips}f, Table~\ref{tab-e1}) and moment-of-inertia (Table~\ref{tab-e1}) errors below HIT, constant-density hulls, and anthropometric regression tables \parencite{deleva1996adjustments, dumas2007adjustments}.

\subsection{Bones, Muscles, and Organs on Held-Out CT}

\begin{figure}[!htb]
\centering
\includegraphics[width=0.99\linewidth,height=0.40\textheight,keepaspectratio]{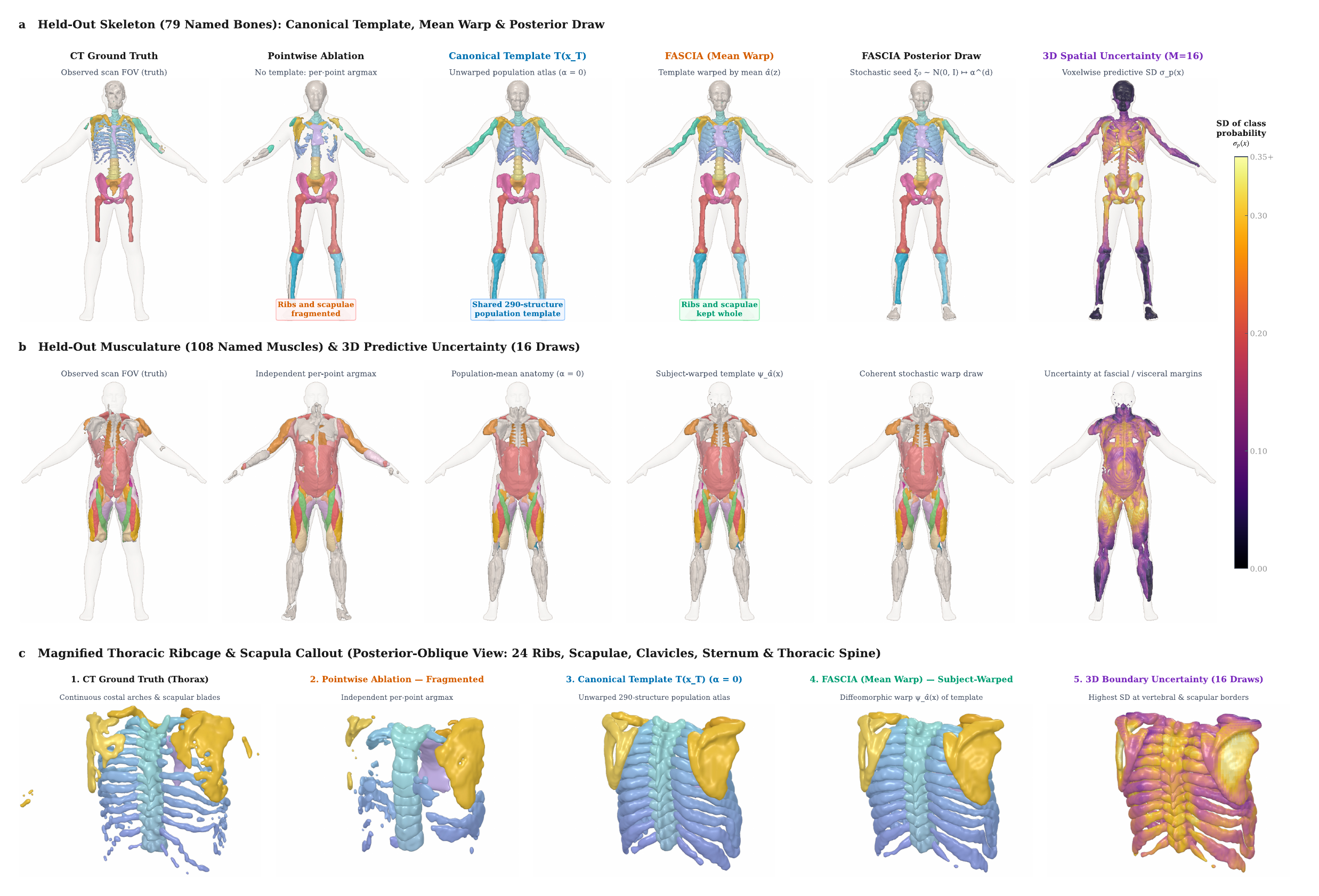}
\caption[]{\textbf{Template-and-warp predictions and posterior samples on held-out CT.} \textbf{(a, b)} Predicted cortical skeleton and musculature on a held-out CT scan comparing ground truth, pointwise ablation, unwarped template ($\alpha = \mathbf{0}$), FASCIA mean warp, one posterior draw (\textbf{Eq.~\ref{eq-flow-draw}}), and 16-draw probability SD. \textbf{(c)} Posterior-oblique thorax showing intact warped ribs and scapulae versus pointwise fragmentation.}
\label{fig-flow-arch}
\end{figure}

On the held-out CT cohort, we compared FASCIA against the identically conditioned pointwise field $f(x_c, z)$, the unwarped template (\texttt{zero warp}), and the per-scan \texttt{oracle warp} (Table~\ref{tab-ct-ablation}). FASCIA outperformed both ablations, showing the value of our diffeomorphic warp model. The gains are largest on thin curved bones that shift across individuals by more than their cortical thickness, such as ribs and scapulae: pointwise classification blurs cross-subject positional uncertainty into disconnected fragments, whereas warping a single topological template keeps nearly all lattice structures and every exported bone mesh intact (Figure~\ref{fig-flow-arch}). Damping canonical bone as interior Dirichlet boundaries in the shape-blendshape solve also cuts deep-bone drift ($6.9 \to 0.12\text{ mm}$, 33 held-out MRIs) without the folding of hard-clamped bone.

\begin{table}[!htb]
\centering\scriptsize\setlength{\tabcolsep}{2.8pt}\renewcommand{\arraystretch}{1.02}
\centering
\caption[]{Held-out CT ablation on the $5\text{ mm}$ canonical lattice ($n=82$ scans). Cells report mean volumetric Dice / single-piece rate (\%; bold marks the best label-blind predictor). Zero warp ($\alpha = \mathbf{0}$) evaluates the unwarped template $T$ (Dice only), subcutaneous fat forms a single outer sleeve (Dice only), and Oracle warp fits each scan's labels directly with $T$ fixed (upper bound).}
\label{tab-ct-ablation}
\begin{tabular}{>{\raggedright\arraybackslash}p{\dimexpr 0.27\linewidth-2\tabcolsep}>{\centering\arraybackslash}p{\dimexpr 0.14\linewidth-2\tabcolsep}>{\centering\arraybackslash}p{\dimexpr 0.20\linewidth-2\tabcolsep}>{\centering\arraybackslash}p{\dimexpr 0.21\linewidth-2\tabcolsep}>{\centering\arraybackslash}p{\dimexpr 0.18\linewidth-2\tabcolsep}}
\toprule
Structure & Zero warp & Pointwise $f(x_c, z)$ & FASCIA & Oracle warp \\
\hline
Named bones (79) & 0.361 & 0.411 / 71\% & \textbf{0.476} / \textbf{93\%} & 0.622 / 95\% \\
$\;\;$Ribs & 0.165 & 0.101 / 29\% & \textbf{0.258} / \textbf{85\%} & 0.418 / 90\% \\
$\;\;$Scapulae & 0.263 & 0.267 / 18\% & \textbf{0.404} / \textbf{100\%} & 0.533 / 100\% \\
Muscles (108) & 0.505 & 0.586 / 85\% & \textbf{0.634} / \textbf{98\%} & 0.785 / 98\% \\
Organs \& vessels & 0.346 & 0.423 / 78\% & \textbf{0.459} / \textbf{81\%} & 0.593 / 81\% \\
Subcutaneous fat & 0.761 & 0.751 & \textbf{0.813} & 0.875 \\
\bottomrule
\end{tabular}
\end{table}

On deep fat compartments and forearm bones---where local compartment thickness or forearm pronation varies independently of smooth whole-body template transport---the pointwise field remains competitive.

\subsection{Uncertainty Quantification and Volumetric Calibration}

Evaluating $T$ at $\hat{\alpha}(z) = \mu_\theta(z)$ gives sharp point predictions, while posterior draws $\alpha^{(d)}$ from \textbf{Eq.~\ref{eq-flow-draw}} deform $T$ diffeomorphically, localizing spatial uncertainty to tissue interfaces (Figure~\ref{fig-flow-arch}a,b) and ranking scan difficulty via $g_\theta(z)$. Following prior work on calibrated uncertainty in markerless biomechanics \parencite{pierzchlewicz_optimizing_2023, pierzchlewicz_platypose_2024, cotton_biomechanical_2025, donahue_calibrated_2026}, we evaluate both voxelwise probability calibration and structure-level interval coverage. At the voxel level, predicted probabilities are reasonably calibrated: modeling both the mean warp and inter-subject warp variability lowers 20-bin Expected Calibration Error (ECE) \parencite{naeini2015obtaining} from 0.097 for the pointwise field to 0.044 across 16 posterior draws (0.051 at the mean warp alone). While voxel probabilities are well calibrated, discrete whole-template draws still exhibit a systematic volume bias: representing thin cortices as connected surfaces on a $5\text{ mm}$ grid dilates enclosed thin-bone meshes ($\sim 1.85\times$, visible where adjacent ribs begin to contact one another in Figure~\ref{fig-flow-arch}c), and because the 32-mode warp shifts boundaries smoothly rather than rescaling each organ, $95\%$ intervals cover $68\text{--}78\%$ of true bone and muscle volumes and bone centroids.

\subsection{Anatomical Plausibility Under Reposing}

Because the volumetric posing map $\Phi$ blends joint transformations smoothly across interior voxels without explicitly constraining bones to move rigidly---whereas exported bone meshes attach rigidly to MHR joints---we test whether reposed volumetric anatomy remains physically plausible without bone stretching, bending, or tissue folding. Across the 33 held-out MRI subjects reposed into 8 diverse postures (264 subject--pose pairs), volumetric bone masks follow rigid skeletal motion closely: lower-limb and forearm bones change length by at most $1.2\%$ (femur/tibia $0.8\%/0.7\%$, radius/ulna $1.1\%/1.2\%$) and bend by $1\text{--}3^\circ$ ($7.0^\circ$ for the ulna). Only the humerus lengthens noticeably across blended shoulder-to-elbow weights ($5.7\%$, reduced by $\delta_s$ from $7.4\%$ for the harmonic rig alone), while folded interior tetrahedra ($0.14\%$ vs. $0.13\%$) and mean regional volume drift ($3.5\%$ vs. $3.1\%$) remain minimal.

\subsection{Driving FASCIA from Images and Video}

\begin{figure}[!htb]
\centering
\includegraphics[width=0.99\linewidth,height=0.31\textheight,keepaspectratio]{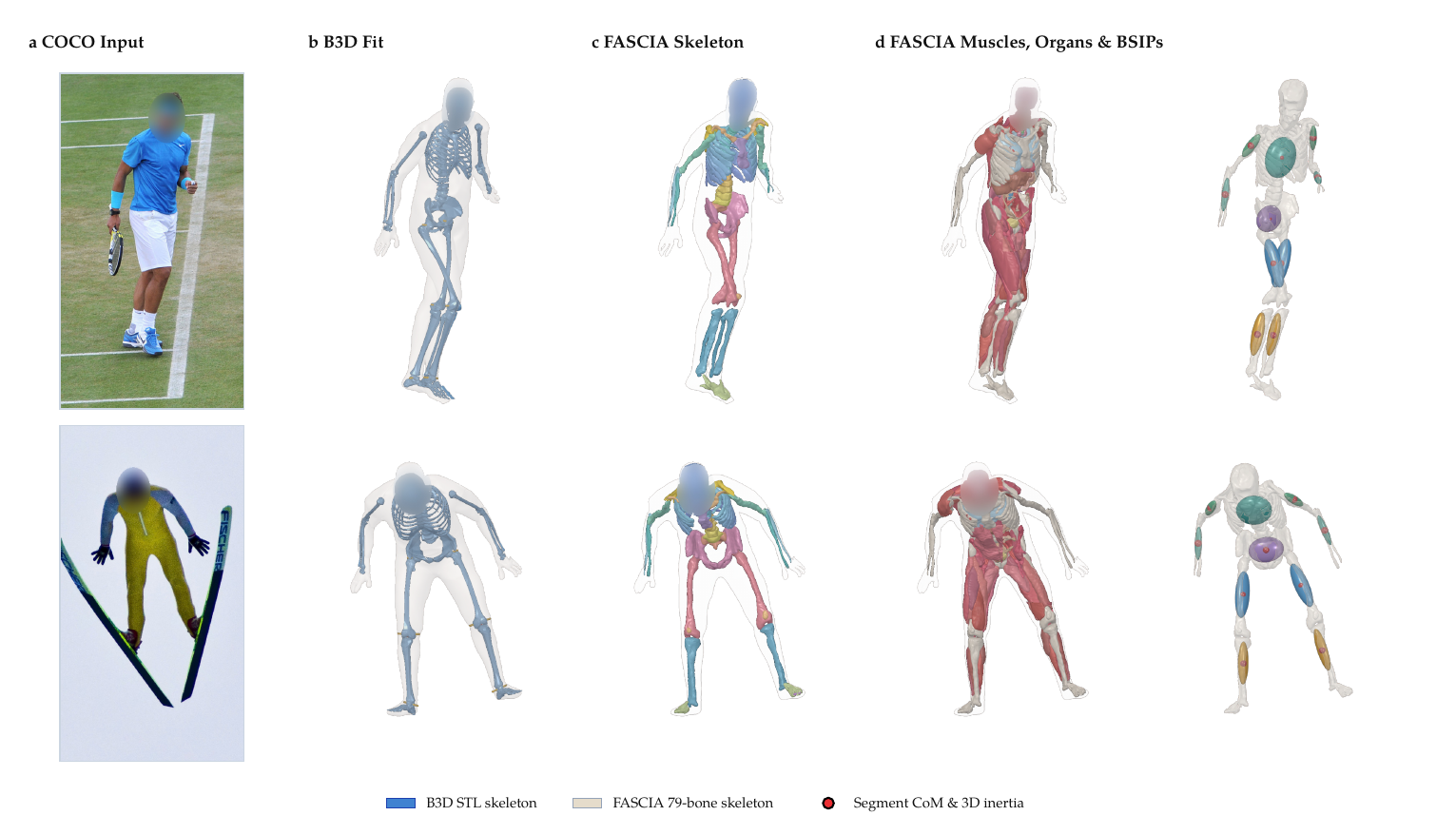}
\caption[]{\textbf{Driving FASCIA from monocular images via B3D.} De-identified COCO-2017 \parencite{lin_microsoft_2014} examples (rows 1--2): \textbf{(a)} Face-blurred RGB inputs. \textbf{(b)} B3D surface fit and height-calibrated biomechanical STL skeleton. \textbf{(c)} Posed FASCIA cortical skeleton. \textbf{(d)} Posed musculature and segment inertia ellipsoids.}
\label{fig-b3d-monocular}
\end{figure}

Driving FASCIA from single de-identified COCO-2017 photographs \parencite{lin_microsoft_2014} via B3D \parencite{cotton_biomechanical_2026} maps image-estimated MHR shape, bone scales, and kinematics directly to intact cortical bones, muscles, organs, and 3D inertia ellipsoids (Figure~\ref{fig-b3d-monocular}); the same path runs frame by frame on monocular video. On an NVIDIA A100 GPU, canonical personalization takes $\sim 34\text{ ms}$ once per subject, and posing takes $\sim 23\text{ ms}$ per frame across the $5\text{ mm}$ lattice or $\sim 10\text{ ms}$ for rigged meshes.

\section{Discussion}

FASCIA predicts fine-grained 3D internal anatomy from MHR shape and skeletal scale and poses it in milliseconds without per-subject optimization. Where thin curved bones shift across people by more than their thickness, pointwise classification fragments them; warping one canonical template via $\psi_\alpha$ morphs bone and soft-tissue geometry while preserving single-piece topology by design. Because two people with the same exterior do not share the same internal anatomy, predicting both the mean warp and its body-specific spread from $z$ yields sharp point estimates alongside complete, coherent posterior draws that quantify spatial and volumetric uncertainty.

Mean overlap alone does not guarantee capturing inter-subject differences: HIT achieves moderate coarse Dice yet collapses adipose and muscle fractions toward the cohort mean. Because MHR, like SMPL, parameterizes only the outer surface and skeleton, recovering inter-subject composition comes from the FASCIA modeling framework, combining forward scan-space supervision and scanner-table modeling $\delta_s$ with a single body-mass measurement to remove the remaining cohort offset. For whole-body fat percentage alone, stature and mass already explain most variance (Table~\ref{tab-e1}); FASCIA's value lies in where that tissue is distributed in 3D, which determines muscle volumes and segment inertia tensors beyond the bulk composition that single-slice opportunistic CT \parencite{marusich_deep_2026}, smartphone photos \parencite{zhou2026photoscan}, and 3D gait \parencite{gabet2026gait} already estimate. Harmonic interior skinning keeps forearm and lower-limb bones within $1.2\%$ of their length under reposing, and prospective DXA validation would test the predicted composition directly.

\subsection{Limitations and Future Directions}

The pointwise field remains competitive on CT deep fat, forearm bones, and total bone volume, and trains directly on MRI labels whereas the canonical template is estimated from CT alone. Because a smooth warp changes compartment volume only through its local Jacobian, estimating a joint CT--MRI template with explicit compartment-volume latents and stronger skeletal geometric priors is more promising than enlarging the warp prior alone. External shape $z$ also does not uniquely determine internal skeletal placement (Table~\ref{tab-ct-ablation}), motivating test-time conditioning on palpable bony landmarks; in addition, structure-level volume intervals remain under-covered. At $5\text{ mm}$ spacing, thin cortices dilate enough for adjacent ribs and vertebrae to contact one another (Figure~\ref{fig-flow-arch}c), and hand bones and toe phalanges remain poorly resolved. Individual arm and forearm muscles are also unparcellated---primarily because MuscleMap \parencite{mckay2024musclemap} does not segment upper-limb muscles, compounded by partial CT arm coverage---and the humerus lengthens modestly under reposing; future work will increase template resolution to resolve individual hand bones and toe phalanges and extend parcellation to upper-limb musculature. In addition, external shape and body mass cannot observe bone mineral density or full visceral fat, our mass-calibrated benchmark uses scan-integrated mass as an upper bound on scale-measured mass, and training covers supine CT and prone MRI in adults without upright organ shift, active muscle bulging, severe sarcopenia, or pediatric anatomy \parencite{bregier2025anny}.

\section*{Acknowledgments}
{\small
AI-assisted coding and research agents, including Claude (Anthropic) and Google Antigravity, were used for code development, running experiments, and assistance with manuscript preparation. The manuscript was extensively reviewed and verified by the authors, who take full responsibility for its content.
\par}

{\small
\printbibliography

@inproceedings{loper_smpl_2015,
  title = {{SMPL}: A Skinned Multi-Person Linear Model},
  author = {Loper, Matthew and Mahmood, Naureen and Romero, Javier and Pons-Moll, Gerard and Black, Michael J.},
  booktitle = {ACM Transactions on Graphics (Proc. SIGGRAPH Asia)},
  volume = {34},
  number = {6},
  pages = {248:1--248:16},
  year = {2015},
  doi = {10.1145/2816795.2818013},
  zotero_key = {ISR3UVZU}
}

@article{ferguson_mhr_2025,
  title = {{MHR}: {Momentum} {Human} {Rig}},
  author = {Ferguson, Aaron and Osman, Ahmed A. A. and Bescos, Berta and Stoll, Carsten and Twigg, Chris and Lassner, Christoph and Otte, David and Vignola, Eric and Prada, Fabian and Bogo, Federica and Santesteban, Igor and Romero, Javier and Zarate, Jenna and Lee, Jeongseok and Park, Jinhyung and Yang, Jinlong and Doublestein, John and Venkateshan, Kishore and Kitani, Kris and Kavan, Ladislav and Farra, Marco Dal and Hu, Matthew and Cioffi, Matthew and Fabris, Michael and Ranieri, Michael and Modarres, Mohammad and Kadlecek, Petr and Khirodkar, Rawal and Abdrashitov, Rinat and Prevost, Romain and Rajbhandari, Roman and Mallet, Ronald and Pearsall, Russel and Kao, Sandy and Kumar, Sanjeev and Parrish, Scott and Yu, Shoou-I. and Saito, Shunsuke and Shiratori, Takaaki and Wang, Te-Li and Tung, Tony and Xu, Yichen and Dong, Yuan and Chen, Yuhua and Xu, Yuanlu and Ye, Yuting and Jiang, Zhongshi},
  journal = {arXiv preprint arXiv:2511.15586},
  year = {2025},
  doi = {10.48550/arXiv.2511.15586},
  url = {http://arxiv.org/abs/2511.15586},
  note = {arXiv:2511.15586},
  zotero_key = {XGSRYYL5}
}

@inproceedings{lin_microsoft_2014,
  title = {Microsoft {COCO}: {Common} {Objects} in {Context}},
  author = {Lin, Tsung-Yi and Maire, Michael and Belongie, Serge and Hays, James and Perona, Pietro and Ramanan, Deva and Dollár, Piotr and Zitnick, C. Lawrence},
  booktitle = {Computer Vision -- ECCV 2014},
  publisher = {Springer International Publishing},
  pages = {740--755},
  year = {2014},
  doi = {10.1007/978-3-319-10602-1_48},
  zotero_key = {DV5QVEMQ}
}

@article{delp_opensim_2007,
  title = {{OpenSim}: Open-Source Software to Create and Analyze Dynamic Simulations of Movement},
  author = {Delp, Scott L. and Anderson, Frank C. and Arnold, Allison S. and Loan, Peter and Habib, Ayman and John, Chand T. and Guendelman, Eran and Thelen, Darryl G.},
  journal = {IEEE Transactions on Biomedical Engineering},
  volume = {54},
  number = {11},
  pages = {1940--1950},
  year = {2007},
  doi = {10.1109/TBME.2007.901024},
  zotero_key = {LC6J7F6X}
}

@inproceedings{keller2023skel,
  title     = {From Skin to Skeleton: Towards Biomechanically Accurate {3D} Digital Humans},
  author    = {Keller, Marilyn and Werling, Keenon and Shin, Soyong and Delp, Scott and
               Pujades, Sergi and Liu, C. Karen and Black, Michael J.},
  booktitle = {ACM Transactions on Graphics (Proc.\ SIGGRAPH Asia)},
  volume    = {42},
  number    = {6},
  pages     = {253:1--253:12},
  month     = dec,
  year      = {2023},
  doi       = {10.1145/3618381},
  zotero_key = {M347TQEP}
}

@inproceedings{keller2024hit,
  title     = {{HIT}: Estimating Internal Human Implicit Tissues from the Body Surface},
  author    = {Keller, Marilyn and Arora, Vaibhav and Dakri, Abdelmouttaleb and Chandhok, Shivam
               and Machann, J{\"u}rgen and Fritsche, Andreas and Black, Michael J. and Pujades, Sergi},
  booktitle = {IEEE/CVF Conference on Computer Vision and Pattern Recognition (CVPR)},
  pages     = {3480--3490},
  month     = jun,
  year      = {2024},
  doi       = {10.1109/CVPR52733.2024.00334},
  zotero_key = {TSCNZU8Q}
}

@inproceedings{alvarado2026soma,
  title     = {{SOMA}: From Surface Observations to Muscle Anatomy},
  author    = {Alvarado, Eduardo and Kim, Emily and Nolte, Gerrit and Runte, Friedemann
               and Botsch, Mario and Habermann, Marc and Theobalt, Christian},
  booktitle = {European Conference on Computer Vision (ECCV)},
  year      = {2026},
  eprint    = {2606.09246},
  archivePrefix = {arXiv},
  primaryClass  = {cs.CV},
  url       = {http://arxiv.org/abs/2606.09246},
  zotero_key = {KFT9MCVM}
}

@article{ashburner2007dartel,
  title   = {A fast diffeomorphic image registration algorithm},
  author  = {Ashburner, John},
  journal = {NeuroImage},
  volume  = {38},
  number  = {1},
  pages   = {95--113},
  year    = {2007},
  doi     = {10.1016/j.neuroimage.2007.07.007},
  zotero_key = {SJBLPXYB}
}

@article{joshi2004unbiased,
  zotero_key = {D49GXQL5},
  title   = {Unbiased diffeomorphic atlas construction for computational anatomy},
  author  = {Joshi, Sarang and Davis, Brad and Jomier, Matthieu and Gerig, Guido},
  journal = {NeuroImage},
  volume  = {23},
  pages   = {S151--S160},
  year    = {2004},
  doi     = {10.1016/j.neuroimage.2004.07.068}
}

@inproceedings{dalca2019conditional,
  zotero_key = {TGENUN3Q},
  title     = {Learning Conditional Deformable Templates with Convolutional Networks},
  author    = {Dalca, Adrian V. and Rakic, Marianne and Guttag, John and Sabuncu, Mert R.},
  booktitle = {Advances in Neural Information Processing Systems (NeurIPS)},
  year      = {2019},
  eprint    = {1908.02738},
  archivePrefix = {arXiv}
}

@article{bregier2025anny,
  title         = {Human Mesh Modeling for {Anny} Body},
  author        = {Br{\'e}gier, Romain and Fiche, Gu{\'e}nol{\'e} and Bravo-S{\'a}nchez, Laura and
                   Lucas, Thomas and Armando, Matthieu and Weinzaepfel, Philippe and Rogez, Gr{\'e}gory
                   and Baradel, Fabien},
  journal       = {arXiv preprint arXiv:2511.03589},
  year          = {2025},
  eprint        = {2511.03589},
  archivePrefix = {arXiv},
  zotero_key = {3BU7SNYG}
}

@inproceedings{keller2022osso,
  title     = {{OSSO}: Obtaining Skeletal Shape from Outside},
  author    = {Keller, Marilyn and Zuffi, Silvia and Black, Michael J. and Pujades, Sergi},
  booktitle = {IEEE/CVF Conference on Computer Vision and Pattern Recognition (CVPR)},
  pages     = {20460--20469},
  year      = {2022},
  doi       = {10.1109/CVPR52688.2022.01984},
  eprint    = {2204.10129},
  archivePrefix = {arXiv},
  zotero_key = {ZHR9AXJ7}
}

@article{shetty2023boss,
  title   = {{BOSS}: Bones, organs and skin shape model},
  author  = {Shetty, Karthik and Birkhold, Annette and Jaganathan, Srikrishna and Strobel, Norbert and
             Egger, Bernhard and Kowarschik, Markus and Maier, Andreas},
  journal = {Computers in Biology and Medicine},
  volume  = {165},
  pages   = {107383},
  year    = {2023},
  doi     = {10.1016/j.compbiomed.2023.107383},
  eprint  = {2303.04923},
  archivePrefix = {arXiv},
  zotero_key = {GMANU78L}
}

@article{wasserthal2023totalsegmentator,
  zotero_key = {RLRITTNW},
  title   = {{TotalSegmentator}: robust segmentation of 104 anatomic structures in {CT} images},
  author  = {Wasserthal, Jakob and Breit, Hanns-Christian and Meyer, Manfred T. and Pradella, Maurice
             and Hinck, Daniel and Sauter, Alexander W. and Heye, Tobias and Boll, Daniel and
             Cyriac, Joshy and Yang, Shan and Bach, Michael and Segeroth, Martin},
  journal = {Radiology: Artificial Intelligence},
  volume  = {5},
  number  = {5},
  pages   = {e230024},
  year    = {2023},
  doi     = {10.1148/ryai.230024},
  eprint  = {2208.05868},
  archivePrefix = {arXiv}
}

@article{sundar2022moose,
  zotero_key = {P75NLQ3K},
  title   = {Fully Automated, Semantic Segmentation of Whole-Body {18F-FDG} {PET/CT} Images Based on
             Data-Centric Artificial Intelligence},
  author  = {Shiyam Sundar, Lalith Kumar and Yu, Josef and Muzik, Otto and Kulterer, Oana C. and
             Fueger, Barbara and Kifjak, Daria and Nakuz, Thomas and Shin, Hyung Min and Sima, Annika
             Katharina and Kitzmantl, Daniela and Badawi, Ramsey D. and Nardo, Lorenzo and
             Cherry, Simon R. and Spencer, Benjamin A. and Hacker, Marcus and Beyer, Thomas},
  journal = {Journal of Nuclear Medicine},
  volume  = {63},
  number  = {12},
  pages   = {1941--1948},
  year    = {2022},
  doi     = {10.2967/jnumed.122.264063},
  note    = {MOOSE}
}

@article{mckay2024musclemap,
  zotero_key = {TME5SFME},
  title   = {{MuscleMap}: An Open-Source, Community-Supported Consortium for Whole-Body Quantitative
             {MRI} of Muscle},
  author  = {McKay, Marnee J. and Weber, Kenneth A. and Wesselink, Evert O. and Smith, Zachary A. and others},
  journal = {Journal of Imaging},
  volume  = {10},
  number  = {11},
  pages   = {262},
  year    = {2024},
  doi     = {10.3390/jimaging10110262}
}

@article{xu2025cads,
  zotero_key = {7TFQIWTN},
  title      = {{CADS}: A Comprehensive Anatomical Dataset and Segmentation for Whole-Body Anatomy
                in Computed Tomography},
  author     = {Xu, Murong and Navarro, Fernando and Fritsak, Maksym and Hamamci, Ibrahim Ethem and others},
  journal    = {Nature Communications},
  year       = {2026},
  doi        = {10.1038/s41467-026-78732-1},
  eprint     = {2507.22953},
  archivePrefix = {arXiv}
}

@article{qiao2024bodycomp,
  zotero_key = {YJX6PMDQ},
  title   = {Prediction of total and regional body composition from {3D} body shape},
  author  = {Qiao, Chexuan and De Lucia Rolfe, Emanuella and Mak, Ethan and Sengupta, Akash and
             Powell, Richard and Watson, Laura P. E. and Heymsfield, Steven B. and Shepherd, John A.
             and Wareham, Nicholas and Brage, Soren and Cipolla, Roberto},
  journal = {npj Digital Medicine},
  volume  = {7},
  number  = {1},
  pages   = {298},
  year    = {2024},
  doi     = {10.1038/s41746-024-01289-0}
}

@article{zhou2026photoscan,
  title         = {Beyond {BMI}: Smartphone Body Composition Phenotyping for Cardiometabolic Risk Assessment},
  author        = {Zhou, Menglian and Charton, Arno and Blanchard, Emily and Cai, Lawrence and Giest, Tracy and Watkins, Herschel and Bouterfa, Mohamed and Wasson, Jackie and Natarajan, Keerthana and Deshpande, Aniket and Zhan, Jiening and Yuen, Shelten and Prieto, Xavi and Shreibati, Jacqueline and Malhotra, Mark and Patel, Shwetak and Sunden, Lindsey and Speed, Cathy and Kokoszka, Alicia and Natarajan, Aravind and Pantelopoulos, Alexandros and Metwally, Ahmed},
  year          = {2026},
  eprint        = {2603.27017},
  archivePrefix = {arXiv},
  doi           = {10.48550/arXiv.2603.27017},
  url           = {http://arxiv.org/abs/2603.27017},
  journal       = {arXiv preprint arXiv:2603.27017},
  zotero_key = {YXZ8Z8KJ}
}

@article{gabet2026gait,
  title         = {A gait foundation model for prediction of multisystem health phenotypes from {3D} skeletal motion},
  author        = {Gabet, Adam and Kohn, Sarah and Lutsker, Guy and Gelman, Shira and Godneva, Anastasia and Sasson, Gil and Zulti, Arad and Krongauz, David and Shaulitch, Rotem and Rotem, Assaf and Doron, Ohad and Brodsky, Yuval and Weinberger, Adina and Segal, Eran},
  journal       = {Nature Medicine},
  year          = {2026},
  pages         = {1--12},
  doi           = {10.1038/s41591-026-04724-9},
  url           = {https://www.nature.com/articles/s41591-026-04724-9},
  zotero_key = {A2JKLAWY}
}

@inproceedings{cotton_biomechanical_2026,
  title     = {Biomechanical {3D} Body: Self-Supervised Distillation of Biomechanical Pose from a {3D} Body Foundation Model},
  author    = {Cotton, R. James and Peiffer, J. D. and Williamson, Lucinda and Leske, John and Pavlakos, Georgios},
  booktitle = {ECCV Workshop on Human Motion Challenges in Real-World and Clinical Settings (MoCha)},
  year      = {2026},
  doi       = {10.48550/arXiv.2608.29928},
  note      = {arXiv:2608.29928},
  zotero_key = {IPR2WE3E}
}

@article{gambietz2026from,
  title   = {From body hulls to musculoskeletal models: Personalized inertial parameter estimation},
  author  = {Gambietz, Markus and Azam, Putri Qistina and Amon, Philipp and Wechsler, Iris and
             Hille, Eva Maria and Menzel, Timo and Ott, Tabea and Botsch, Mario and Braun, Matthias
             and Miehling, J{\"o}rg and McMahon, Katie L. and Koelewijn, Anne D.},
  journal = {PLOS ONE},
  volume  = {21},
  number  = {6},
  pages   = {e0349886},
  year    = {2026},
  doi     = {10.1371/journal.pone.0349886},
  zotero_key = {2ZBVQ3AT}
}

@article{hinge2026copenhagen,
  title   = {A multimodal total-body dynamic {[18F]FDG} {PET/CT/MRI} dataset of 100 healthy humans},
  author  = {Hinge, Christian and H{\o}i-Hansen, Frederikke E. and Hansen, Rasmus H. and
             Korsholm, Kirsten and Johannesen, Helle H. and Albrecht-Beste, Elisabeth and
             Saxtoft, Eunice S. and Jensen, Vibeke and Schulze, Christina and
             Mikkelsen, Christina and Ostrowski, Sisse R. and Nissen, Janna and
             Pernet, Cyril and Fischer, Barbara M. and Ladefoged, Claes N. and
             Andersen, Flemming L.},
  journal = {Scientific Data},
  year    = {2026},
  doi     = {10.1038/s41597-026-08157-4},
  zotero_key = {T97ZB8Y2}
}

@article{akhundov2022subject,
  title   = {Is subject-specific musculoskeletal modelling worth the extra effort or is generic
             modelling worth the shortcut?},
  author  = {Akhundov, Riad and Saxby, David J. and Diamond, Laura E. and Edwards, Suzi and
             Clausen, Phil and Dooley, Katherine and Blyton, Sarah and Snodgrass, Suzanne J. and
             Carlier, Aur{\'e}lie},
  journal = {PLOS ONE},
  volume  = {17},
  number  = {1},
  pages   = {e0262936},
  year    = {2022},
  doi     = {10.1371/journal.pone.0262936},
  zotero_key = {7XUQ4FSI}
}

@incollection{dumas2018estimation,
  title     = {Estimation of the Body Segment Inertial Parameters for the Rigid Body Biomechanical
               Models Used in Motion Analysis},
  author    = {Dumas, Rapha{\"e}l and Wojtusch, Janis},
  booktitle = {Handbook of Human Motion},
  pages     = {47--77},
  year      = {2018},
  doi       = {10.1007/978-3-319-14418-4_147},
  zotero_key = {28ACF9SL}
}

@article{werling2023addbiomechanics,
  title   = {{AddBiomechanics}: Automating model scaling, inverse kinematics, and inverse dynamics
             from human motion data through sequential optimization},
  author  = {Werling, Keenon and Bianco, Nicholas A. and Raitor, Michael and Stingel, Jon and
             Hicks, Jennifer L. and Collins, Steven H. and Delp, Scott L. and Liu, C. Karen},
  journal = {PLOS ONE},
  volume  = {18},
  number  = {11},
  pages   = {e0295152},
  year    = {2023},
  doi     = {10.1371/journal.pone.0295152},
  zotero_key = {MD4SXIKR}
}

@article{jimenezdeltoro2016visceral,
  zotero_key = {8ERWZE5B},
  title   = {Cloud-Based Evaluation of Anatomical Structure Segmentation and Landmark Detection
             Algorithms: {VISCERAL} Anatomy Benchmarks},
  author  = {Jimenez-del-Toro, Oscar and M{\"u}ller, Henning and Krenn, Markus and Gruenberg, Katharina
             and Taha, Abdel Aziz and Winterstein, Marianne and Eggel, Ivan and Foncubierta-Rodríguez,
             Antonio and Goksel, Orcun and Jakab, Andr{\'a}s and Kontokotsios, Georgios and
             Langs, Georg and Menze, Bjoern H. and Salas Fernandez, Tom{\`a}s and Schaer, Roger and
             Walleyo, Anna and Weber, Marc-Andr{\'e} and Dicente Cid, Yashin and Gass, Tobias and
             Heinrich, Mattias and Jia, Fucang and Kahl, Fredrik and Kechichian, Razmig and
             Mai, Dominic and Spanier, Assaf B. and Vincent, Graham and Wang, Chunliang and
             Wyeth, Daniel and Hanbury, Allan},
  journal = {IEEE Transactions on Medical Imaging},
  volume  = {35},
  number  = {11},
  pages   = {2459--2475},
  year    = {2016},
  doi     = {10.1109/TMI.2016.2578680}
}

@article{fischer2023vsd,
  zotero_key = {QLDF4CWZ},
  title   = {Database of segmentations and surface models of bones of the entire lower body created from cadaver {CT} scans},
  author  = {Fischer, Maximilian C. M.},
  journal = {Scientific Data},
  volume  = {10},
  pages   = {763},
  year    = {2023},
  doi     = {10.1038/s41597-023-02669-z}
}

@article{peiffer_portable_2026,
  title   = {Portable biomechanics laboratory enables clinically accessible movement analysis from a handheld smartphone},
  author  = {Peiffer, J. D. and Shah, Kunal and Djuraskovic, Irina and Anarwala, Shawana and Abdou, Kayan and Patel, Rujvee and Jayabalan, Prakash and Pennicooke, Brenton and Cotton, R. James},
  journal = {npj Digital Medicine},
  volume  = {9},
  number  = {1},
  pages   = {692},
  year    = {2026},
  doi     = {10.1038/s41746-026-02825-w},
  zotero_key = {PGV2WTKH}
}

@article{donahue_calibrated_2026,
  author  = {Donahue, Seth and Djuraskovic, Irina and Shah, Kunal and Sinz, Fabian and Chafetz, Ross and Cotton, R. James},
  title   = {Calibrated Uncertainty for Trustworthy Clinical Gait Analysis Using Probabilistic Multiview Markerless Motion Capture},
  journal = {IEEE Transactions on Biomedical Engineering},
  volume  = {73},
  number  = {8},
  pages   = {2889--2898},
  year    = {2026},
  doi     = {10.1109/TBME.2026.3691128},
  note    = {arXiv:2601.22412},
  zotero_key = {ESYJ5B34}
}

@inproceedings{cotton_optimizing_2023,
  title     = {Optimizing {Trajectories} and {Inverse} {Kinematics} for {Biomechanical} {Analysis} of {Markerless} {Motion} {Capture} {Data}},
  author    = {Cotton, R. James and DeLillo, Allison and Cimorelli, Anthony and Shah, Kunal and Peiffer, J. D. and Anarwala, Shawana and Abdou, Kayan and Karakostas, Tasos},
  booktitle = {2023 International Conference on Rehabilitation Robotics (ICORR)},
  pages     = {1--6},
  year      = {2023},
  publisher = {IEEE},
  doi       = {10.1109/ICORR58425.2023.10304683},
  zotero_key = {BYMFMV2C}
}

@article{cimorelli_validation_2024,
  title      = {Validation of portable in-clinic video-based gait analysis for prosthesis users},
  author     = {Cimorelli, Anthony and Patel, Ankit and Karakostas, Tasos and Cotton, R. James},
  journal    = {Scientific Reports},
  volume     = {14},
  number     = {1},
  pages      = {3840},
  year       = {2024},
  doi        = {10.1038/s41598-024-53217-7},
  zotero_key = {8TGRW2I6}
}

@article{uhlrich2023opencap,
  title     = {{OpenCap}: Human movement dynamics from smartphone videos},
  author    = {Uhlrich, Scott D and Falisse, Antoine and Kidzi{\'n}ski, {\L}ukasz and Muccini, Julie and Ko, Michael and Chaudhari, Akshay S and Hicks, Jennifer L and Delp, Scott L},
  journal   = {PLOS Computational Biology},
  volume    = {19},
  number    = {10},
  pages     = {e1011462},
  year      = {2023},
  publisher = {Public Library of Science},
  doi       = {10.1371/journal.pcbi.1011462},
  zotero_key = {BA2BHW8L}
}

@article{kanko_concurrent_2021,
  title     = {Concurrent assessment of gait kinematics using marker-based and markerless motion capture},
  author    = {Kanko, Robert M. and Laende, Elise K. and Davis, Elysia M. and Selbie, W. Scott and Deluzio, Kevin J.},
  journal   = {Journal of Biomechanics},
  volume    = {127},
  pages     = {110665},
  year      = {2021},
  doi       = {10.1016/j.jbiomech.2021.110665},
  publisher = {Elsevier},
  zotero_key = {ATHCTVW4}
}

@article{dumas2007adjustments,
  zotero_key = {YX9VXLL3},
  title   = {Adjustments to {McConville} et al. and {Young} et al. body segment inertial parameters},
  author  = {Dumas, Rapha{\"e}l and Ch{\`e}ze, Laurence and Verriest, Jean-Pierre},
  journal = {Journal of Biomechanics},
  volume  = {40},
  number  = {3},
  pages   = {543--553},
  year    = {2007},
  doi     = {10.1016/j.jbiomech.2006.02.013}
}

@article{deleva1996adjustments,
  title   = {Adjustments to {Zatsiorsky-Seluyanov}'s segment inertia parameters},
  author  = {de Leva, Paolo},
  journal = {Journal of Biomechanics},
  volume  = {29},
  number  = {9},
  pages   = {1223--1230},
  year    = {1996},
  doi     = {10.1016/0021-9290(95)00178-6},
  zotero_key = {WKX7698H}
}

@article{gutschmayer2025quadrahc,
  title   = {Whole-Body [{18F}]{FDG}-{PET}/{CT} Imaging of Healthy Controls: Test/Retest Data for Systemic, Multi-Organ Analysis},
  author  = {Gutschmayer, Sebastian and Yu, Josef and Geist, Barbara Katharina and {\"O}zer, {\"O}yk{\"u} and Reiterits, Bettina and Ferrara, Daria and Pires, Manuel and Rausch, Ivo and Ibeschitz, Harald and Karanikas, Georgios and Sundar, Lalith Kumar Shiyam and Nics, Lukas and Chalampalakis, Zacharias and Muin, Dina and Langsteger, Werner and Hacker, Marcus and Beyer, Thomas},
  journal = {Scientific Data},
  volume  = {12},
  number  = {1},
  pages   = {1707},
  year    = {2025},
  doi     = {10.1038/s41597-025-05997-4},
  zotero_key = {ENSAVZSV}
}

@article{pierzchlewicz_platypose_2024,
  title      = {Platypose: Calibrated Zero-Shot Multi-Hypothesis {3D} Human Motion Estimation},
  author     = {Pierzchlewicz, Pawe{\l} A. and Silva, Caio O. da and Cotton, R. James and Sinz, Fabian H.},
  journal    = {arXiv preprint arXiv:2403.06164},
  year       = {2024},
  doi        = {10.48550/arXiv.2403.06164},
  url        = {https://arxiv.org/abs/2403.06164},
  zotero_key = {35LAPNK8}
}

@inproceedings{pierzchlewicz_optimizing_2023,
  title      = {Optimizing {MPJPE} promotes miscalibration in multi-hypothesis human pose lifting},
  author     = {Pierzchlewicz, Pawe{\l} A. and Bashiri, Mohammad and Cotton, R. James and Sinz, Fabian H.},
  booktitle  = {ICLR 2023 Affinity Workshop},
  year       = {2023},
  url        = {https://openreview.net/forum?id=B5riBS9HZGn},
  zotero_key = {JPTQHBFC}
}

@inproceedings{cotton_biomechanical_2025,
  title      = {Biomechanical Reconstruction with Confidence Intervals from Multiview Markerless Motion Capture},
  author     = {Cotton, R. James and Sinz, Fabian H.},
  booktitle  = {2025 47th Annual International Conference of the IEEE Engineering in Medicine and Biology Society (EMBC)},
  pages      = {1--7},
  year       = {2025},
  doi        = {10.1109/EMBC58623.2025.11253412},
  zotero_key = {PXQCMRJA}
}

@article{gneiting2007strictly,
	title = {Strictly Proper Scoring Rules, Prediction, and Estimation},
	author = {Gneiting, Tilmann and Raftery, Adrian E.},
	journal = {Journal of the American Statistical Association},
	volume = {102},
	number = {477},
	pages = {359--378},
	year = {2007},
	doi = {10.1198/016214506000001437},
  zotero_key = {EAL6THRG}
}

@inproceedings{kingma2014autoencoding,
  title      = {Auto-Encoding Variational {Bayes}},
  author     = {Kingma, Diederik P. and Welling, Max},
  booktitle  = {2nd International Conference on Learning Representations (ICLR)},
  year       = {2014},
  zotero_key = {SQAN9J9C}
}

@inproceedings{naeini2015obtaining,
  title      = {Obtaining Well Calibrated Probabilities Using {Bayesian} Binning},
  author     = {Pakdaman Naeini, Mahdi and Cooper, Gregory F. and Hauskrecht, Milos},
  booktitle  = {Proceedings of the Twenty-Ninth AAAI Conference on Artificial Intelligence},
  pages      = {2901--2907},
  year       = {2015},
  doi        = {10.1609/aaai.v29i1.9602},
  zotero_key = {C2TFJL3B}
}

@article{marusich_deep_2026,
  title = {Deep learning reveals relationships between muscle and adipose tissue and their associations with age, body mass index, and low back pain: {A} cross-sectional study in 10,835 subjects},
  author = {Marusich, Kathryn R. and Gatti, Anthony A. and Blankemeier, Louis and Boutin, Robert D. and Smuck, Matthew and Gold, Garry E. and Delp, Scott L. and Chaudhari, Akshay S.},
  journal = {PLOS Digital Health},
  year = {2026},
  volume = {5},
  number = {11},
  pages = {e0001788},
  doi = {10.1371/journal.pdig.0001788},
  zotero_key = {BERTR8K2}
}
}

\end{document}